\documentclass[conference]{IEEEtran}
\IEEEoverridecommandlockouts
\usepackage{amsmath,amssymb,amsfonts}
\usepackage{algorithmic}
\usepackage{balance}
\usepackage{booktabs}
\usepackage{listings}
\usepackage{multirow}
\usepackage{graphicx}
\usepackage{pifont}
\usepackage{textcomp}
\usepackage{xcolor}
\usepackage[separate-uncertainty=true]{siunitx}
\usepackage[natbib=true]{biblatex}

\bibliography{paper}

\newcommand{\cmark}{\ding{51}}
\newcommand{\xmark}{\ding{55}}

\newcommand{\funding}{%
  This work was funded by the EPSRC, grants EP/V052241/1 and EP/S030964/1; and Intel Corporation. BM is also funded by the Leverhulme Trust and The Alan Turing Institute. 
  Compute time was provided through Gauss Centre for Supercomputing application number 21018 and EPSRC (grant number EP/T022205/1) and local GPU hardware was provided by an NVIDIA hardware grant award.
}
\begin{document}

\title{A Complete Pipeline for deploying SNNs with Synaptic Delays on Loihi 2}
%\title{Training Delay-Enhanced SNNs on GPUs for Neuromorphic Deployment}
%\title{Energy-Efficient Keyword Recognition with Delay-Enhanced SNNs on Neuromorphic Hardware}
%\title{Delay-Enhanced SNNs from GPU Training to Neuromorphic Execution}
\author{%
  \IEEEauthorblockN{%
    Balazs Meszaros\IEEEauthorrefmark{1}, 
    James~C.~Knight\IEEEauthorrefmark{1}, Jonathan Timcheck\IEEEauthorrefmark{2},
    and Thomas Nowotny\IEEEauthorrefmark{1}}
  \IEEEauthorblockA{%
    \IEEEauthorrefmark{1}School of Engineering and Informatics,
                    University of Sussex, Brighton, BN1 9QJ, UK}
    \IEEEauthorrefmark{2}Intel Labs, Intel Corporation, Santa Clara, CA, 95054, USA\\
                    Email: \{bm452, j.c.knight, t.nowotny\}@sussex.ac.uk, jonathan.timcheck@intel.com   
  \thanks{\funding}
}

\maketitle

\begin{abstract}
Spiking Neural Networks are attracting increased attention as a more energy-efficient alternative to traditional Artificial Neural Networks for edge computing. Neuromorphic computing can significantly reduce energy requirements. Here, we present a complete pipeline: efficient event-based training of SNNs with synaptic delays on GPUs and deployment on Intel's Loihi 2 neuromorphic chip. We evaluate our approach on keyword recognition tasks using the Spiking Heidelberg Digits and Spiking Speech Commands datasets, demonstrating that our algorithm can enhance classification accuracy compared to architectures without delays. Our benchmarking indicates almost no accuracy loss between GPU and Loihi 2 implementations, while classification on Loihi 2 is up to 18× faster and uses 250× less energy than on an NVIDIA Jetson Orin Nano.
\end{abstract}


\section{Introduction}
Voice-controlled technologies have become ubiquitous in our daily lives, with smart speakers, mobile devices, laptops, and wearables all offering voice-control capabilities. Keyword recognition is fundamental to these technologies, enabling devices to recognize specific commands that trigger further actions. While many current implementations rely on cloud processing for more complex speech recognition tasks, this approach introduces latency, requires stable internet connectivity, and raises privacy concerns.

Neuromorphic computing offers a promising solution to these challenges by enabling more efficient edge processing. Neuromorphic systems are brain-inspired that typically implement `spiking' neurons which communicate through events that are sparse in space and time, making them particularly well-suited for temporal tasks like keyword recognition. In recent years, numerous neuromorphic platforms have emerged from academic institutions~\citep{pehle2022brainscales,d2024denram}, startups and industry leaders~\citep{davies2021advancing}, demonstrating significant reductions in computation time and energy consumption for spatiotemporal tasks.

Despite these hardware advances, developing algorithms that fully leverage the capabilities of neuromorphic systems remains challenging. A key limitation has been the lack of support for end-to-end learning directly on neuromorphic hardware. Previous work has shown promising results using recurrent SNNs trained with EventProp~\citep{wunderlich2021event,nowotny2025loss} for keyword spotting tasks and deployed on Loihi 2~\citep{shoesmith2025eventprop}, demonstrating substantial improvements in speed and energy efficiency compared to conventional hardware.

Our work builds upon this foundation by specifically addressing the critical role of synaptic delays in SNNs. In biological neural networks, these delays arise naturally from spatial structure and contribute significantly to learning and coincidence detection~\citep{bengtsson2005extensive,seidl2010mechanisms}. From a computational perspective, networks with adjustable delays can compute a richer class of functions than those with only adjustable weights~\citep{izhikevich2006polychronization,maass1999complexity}, and achieve state-of-the-art solutions for keyword recognition~\citep{hammouamrilearning}. Importantly, neuromorphic systems like Loihi are designed to accommodate such delays efficiently~\citep{davies2021advancing}. Although the dilated convolutions-based delay learning algorithm is powerful~\citep{hammouamrilearning}, it does not exploit sparsity in SNNs. ~\citet{goltz2024delgrad} were the first to provide an efficient learning rule for delays, relying on exact gradients. Later, we introduced a similar method based on the EventProp~\citep{wunderlich2021event} formalism~\citep {meszaros2025efficient} and showed that exact gradients can be used to learn delays in tasks where multiple spikes, multiple layers and recurrent architectures are necessary. We subsequently implemented this extension in mlGeNN~\citep{turner2022mlgenn,knight2023easy} -- a user-friendly spike-based ML library built on the GPU-optimized GeNN simulator~\citep{yavuz2016genn,knight2021pygenn}.

\begin{figure}
    \centering
    \includegraphics[width=1\linewidth]{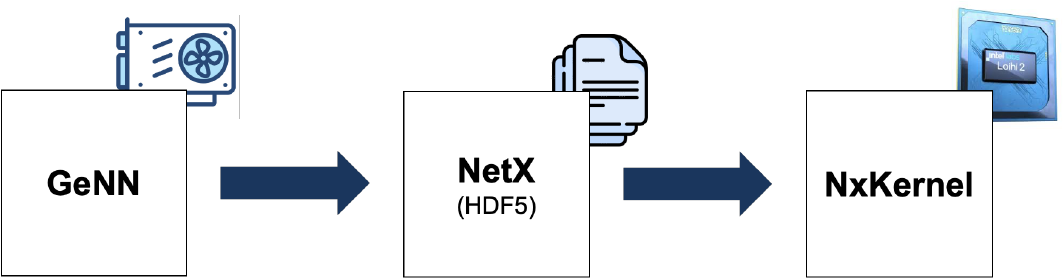}
    \caption{We train models with GeNN and export them in the NetX format for inference on Loihi.}
    \label{fig:gpu_to_loihi}
\end{figure}

In this paper, we present a complete pipeline for training SNNs with learnable delays using EventProp in our mlGeNN framework and deploying them on the Loihi 2 neuromorphic system (see Figure \ref{fig:gpu_to_loihi}).
To our knowledge, this is the first demonstration of deploying SNN models with trained delays on neuromorphic hardware for keyword recognition tasks, representing an important step toward practical, energy-efficient edge computing solutions for speech recognition applications.

\section{Methods}
All code used for this work is publicly available at \url{https://github.com/mbalazs98/deventprop/}, aside from the NxKernel Loihi 2 implementation, which can be made available to members of the Intel Neuromorphic Research Community upon request. 
GeNN, mlGeNN and the mlGeNN to Network Exchange (NetX) converter are available at \url{https://github.com/genn-team/}.

\subsection{Loihi 2}
%\href{https://arxiv.org/pdf/2310.03251}{https://arxiv.org/pdf/2310.03251} }
Intel's Loihi 2 is a fully asynchronous, scalable neuromorphic architecture featuring 120 neuromorphic cores that execute neuronal dynamics in parallel. It offers programmable neuron models with a rich microcode instruction set, supports graded spikes with integer payloads, includes six embedded processor cores, and provides a 10 Gbps Ethernet spike I/O interface. Particularly relevant to this works, it also implements synaptic delays of up to 62 timesteps.

\subsection{mlGeNN}
mlGeNN~\citep{turner2022mlgenn,knight2023easy} is a spike-based ML library built on the GPU-optimised GeNN simulator~\citep{yavuz2016genn,knight2021pygenn} with support for training networks using the EventProp~\citep{wunderlich2021event,nowotny2025loss} learning rule.
Here, we use EventProp, which is a form of event-based Backpropagation Through Time~(BPTT) using exact gradients.
In EventProp, neuron states only need to be stored between the forward and backward passes at spike times, which massively reduces memory requirements compared to other implementations of BPTT.
Furthermore, the EventProp backward pass has very similar computational properties to the forward pass of an SNN, meaning that it can be implemented very efficiently in GeNN.
\setlength{\tabcolsep}{2pt} % Default value: 6pt
\begin{table}
\caption{Test accuracy of models trained on SHD and SSC %and evaluated using mlGeNN using \SI{32}{\bit} floating point as well as on Loihi~2 using the NxKernel API. 
%For mlGeNN results, bar heights represent mean accuracy and error bars the standard deviations, each calculated across 5 models trained with different random seeds. On Loihi~2, bar heights represent the accuracy of a single model.
}
  \centering
  \begin{tabular}{r r S S S S}
    \toprule
        {Dataset} & {Architecture} & \multicolumn{2}{c}{mlGeNN accuracy [\%]} & \multicolumn{2}{c}{Loihi accuracy [\%]}\\
        & & {(no delay)} & {(delay)} & {(no delay)} & {(delay)} \\
    \midrule
        \multirow{2}{*}{SHD} & Feedforward & 72.1(2.6) & 86.9(2.3) & 67.9 & 88.0 \\
         & Recurrent & 87.0(2.1) & 88.8(2.1) & 89.5 & 90.9 \\
         \multirow{2}{*}{SSC} & Feedforward & 44.5(1.1) & 71.4(1.0) & 42.6 & 69.8 \\
         & Recurrent & 63.7(1.0) & 65.3(5.7) & 62.5 & 67.8 \\
    \bottomrule
  \end{tabular}
  \label{tab:accuracy}
\end{table}
\subsection{Datasets}
We used two keyword spotting datasets developed by \citet{cramer2020heidelberg}. The Spiking Heidelberg Digits (SHD) consist of \num{10420} spoken digits (0-9) in German and English, spoken by 12 speakers. The Spiking Speech Commands (SSC) consist of \num{105829} samples, from Google Speech Commands, featuring 35 words spoken by a larger speaker pool.
%\begin{figure}
%    \centering
%    \includegraphics{accuracy.pdf}
%    \caption{Test accuracy of models trained on SHD and SSC and evaluated using mlGeNN using \SI{32}{\bit} floating point as well as on Loihi~2 using the NxKernel API. For mlGeNN results, bar heights represent mean accuracy and error bars the standard deviations, each calculated across 5 models trained with different random seeds. On Loihi~2, bar heights represent the accuracy of a single model.}
%    \label{fig:accuracy}
%\end{figure}
The audio data from both datasets was transformed into spikes across 700 input channels using an artificial cochlea model.

\subsection{Spiking Neural Networks}
\label{sec:snn}
~\citet{nowotny2025loss} extended EventProp~\citep{wunderlich2021event} to be suitable for more complex tasks, and implemented the extensions in mlGeNN. More recently, delay learning was derived and implemented in the framework~\citep{meszaros2025efficient}. 
We trained both single-layer recurrent and 2-layer feedforward networks of Leaky Integrate-and-Fire (LIF) neurons and exponential synapses with and without delays. Output neurons are non-firing leaky integrators, and the classification decision is based on the maximum of the exponentially weighted voltage integrals of the output neurons~\cite{nowotny2025loss}.  For SHD, we used 512 hidden neurons, while for SSC, we used 1024 hidden neurons. We conducted hyperparameter searches over timestep sizes (\SI{1}-\SI{2}{\milli\second}), for various delay initialisation ranges from a uniform distribution and regularisation strengths, with the target firing rate of hidden neurons being set at \SI{14}{\hertz}. Given the lack of an explicit validation set for SHD, we conducted cross-validation on the training set. For SSC we chose our model based on results obtained on the validation set. We report our best-performing models for recurrent and feedforward networks, with and without delays.
\begin{table*}
\caption{Spiking Heidelberg Digits classification performance comparison.}
  \centering
  \begin{tabular}{r r r r S S S S}
    \toprule
        {Architecture} & {Delay} & {Weight} & {Hardware} & \multicolumn{4}{c}{Hardware inference cost per sample}\\
        & & {precision} & & \multicolumn{2}{c}{Energy}  & {Latency} & {EDP}\\
         & & &  & {Total [\si{\milli\joule}]} & {Dynamic [\si{\milli\joule}]} & {[\si{\milli\second}]} & {[\si{\micro\joule\times\second}]}\\
    \midrule
        \multirow{6}{*}{Feedforward} & \cmark & fp32 & Jetson Orin Nano GPU (batch=1)$^\ddagger$ & 93.7 & 30.0 & 27.5 & 2581.9 \\
         & \xmark & fp32 & Jetson Orin Nano GPU (batch=1)$^\ddagger$ & 85.4 & 25.8 & 26.0 & 2219.6 \\
         & \cmark & fp32 & Jetson Orin Nano GPU (batch=4)$^\ddagger$ & 50.6 & 22.6 & 48.5 & 2454.8 \\
         & \xmark & fp32 & Jetson Orin Nano GPU (batch=128)$^\ddagger$ & 8.3 &  4.9 & 178.8 & 1476.9 \\
         & \cmark & int8 & Loihi 2$^\dagger$ & 0.46 & 0.28 & 1.46 &  0.67 \\
         & \xmark & int8 & Loihi 2$^\dagger$ & 0.42 & 0.25 & 1.47 & 0.62 \\
     \midrule 
        \multirow{6}{*}{Recurrent} & \cmark & fp32 & Jetson Orin Nano GPU (batch=1)$^\ddagger$ & 93.3 & 29.0 & 27.9 & 2599.0 \\
         & \xmark & fp32 & Jetson Orin Nano GPU (batch=1)$^\ddagger$ & 85.0 & 25.3 & 26.0 & 2214.8 \\
         & \cmark & fp32 & Jetson Orin Nano GPU (batch=4)$^\ddagger$ & 41.9 & 16.5 & 43.6 & 1827.6 \\
         & \xmark & fp32 & Jetson Orin Nano GPU (batch=128)$^\ddagger$ & 6.5 & 3.7 & 156.4 & 1018.2 \\
         & \cmark & int8 & Loihi 2$^\dagger$ & 0.36 & 0.21 & 1.54 & 0.56 \\
         & \xmark & int8 & Loihi 2$^\dagger$ & 0.33 & 0.19 & 1.49 & 0.50 \\
         % from NICE 2025 & \xmark & int8 & Loihi 2$^\dagger$ & 0.27 & 0.15 & 2.37 & 0.63 \\
     %\midrule
     %    \multirow{6}{*}{SSC FF} & \cmark & fp32 & Jetson Orin Nano GPU (batch=1)$^\ddagger$ &  &  &  &  \\
     %    & \xmark & fp32 & Jetson Orin Nano GPU (batch=1)$^\ddagger$ &  &  & &  \\
     %    & \cmark & fp32 & Jetson Orin Nano GPU (batch=64)$^\ddagger$ &  &  & &  \\
     %    & \xmark & fp32 & Jetson Orin Nano GPU (batch=64)$^\ddagger$ &  &  &  &  \\
     %    & \cmark & int8 & Loihi 2$^\dagger$ &  &  &  &  \\
     %    & \xmark & int8 & Loihi 2$^\dagger$ &  &  & & \\
     %\midrule
     %    \multirow{6}{*}{SSC RC} & \cmark & fp32 & Jetson Orin Nano GPU (batch=1)$^\ddagger$ &  &  &  &  \\
     %    & \xmark & fp32 & Jetson Orin Nano GPU (batch=1)$^\ddagger$ & 92.0 & 29.1 & 27.4 & 2522.5 \\
     %    & \cmark & fp32 & Jetson Orin Nano GPU (batch=64)$^\ddagger$ &  &  &  &  \\
     %    & \xmark & fp32 & Jetson Orin Nano GPU (batch=64)$^\ddagger$ & 15.5 & 9.7 & 161.2 & 2506.6 \\
     %    & \cmark & int8 & Loihi 2$^\dagger$ &  &  &  &  \\
     %    & \xmark & int8 & Loihi 2$^\dagger$ &  &  &  &  \\
        \bottomrule \\
  \end{tabular}
  \centering
          \vskip 0.01em 

  \begin{minipage}{0.78\textwidth}
      $^\dagger$Loihi 2 workloads were characterized on an Oheo Gulch system with N3C2-revision Loihi 2 chips running on an unreleased patch of NxCore 2.5.10 and
alpha version of the NxKernel API with on-chip IO unthrottled sequencing of inputs.
 
 $^\ddagger$Jetson workloads were characterized on an NVIDIA Jetson Orin Nano 8GB 15W TDP running Jetpack 5.1.2, GeNN 5.1.0, mlGeNN 2.3.0 and mlGeNN NetX 0.2.0; energy values include CPU\_GPU\_CV and SOC components as reported by jtop.
 
 $^*$Performance results are based on testing as of April 2025 and may not reflect all publicly available security updates; results may vary.
\end{minipage}
  \label{tab:performance}
\end{table*}

\subsection{mlGeNN to NetX conversion} 
\label{sec:ml_genn_netx}

\citet{shoesmith2025eventprop} introduced a pipeline for exporting trained models from mlGeNN to Network exchange (NetX) -- an HDF5-based network model exchange format designed for importing models from other
frameworks into the Loihi ecosystem. We extended the exporter to support networks with delay, ensuring that delays are clamped at below 62 timesteps to suit the requirements of Loihi 2.

\section{Results}
\subsection{EventProp enables faster training of models with delay using less memory}
Prior work benchmarked mlGeNN-based EventProp delay learning \citep{meszaros2025efficient} against the Dilated Convolution implementation provided by \citet{hammouamrilearning}. In those benchmarks, feedforward models with 2 hidden layers were used since the Dilated Convolution method does not support recurrent delays. Even with the 62 delay slots used here, mlGeNN requires only half the memory compared to the Dilated Convolution approach. Similarly, mlGeNN demonstrates superior computational performance, running up to 10 times faster than Dilated Convolutions.

\subsection{Delays provide significant accuracy improvements on Loihi}
Table \ref{tab:accuracy} compares the test accuracies of the models described in section~\ref{sec:snn}, 
simulated with \SI{32}{\bit} floating point weights using mlGeNN and with quantized \SI{8}{\bit} integer weights on Loihi 2. Note that we report GPU results averaged over 5 runs with different seeds and, for Loihi, we report the results of one seed.
NetX models were deployed using NxKernel, an intermediate-level neuromorphic programming interface available to members of the Intel Neuromorphic Research Community.

While adding connection delays only results in modest benefits for recurrent architectures (1.0\% improvement for SHD, 1.6\% for SSC), feedforward architectures showed significant improvements (13.3\% improvement for SHD, 26.9\% for SSC), achieving the highest accuracy on SSC, and similar results to the recurrent architecture on SHD. 

In our prior work~\citep{meszaros2025efficient}, we trained models with no constraint on the maximum number of delay steps. On both SHD and SSC these unconstrained models outperformed the models presented here -- where the maximum delay is limited 62 timesteps -- by approximately 5\%. Given these results, \SI{62}{\milli\second} might be a heavy limitation for tasks such as keyword recognition.

\subsection{Loihi~2 delivers low-energy, low-latency inference}
Table~\ref{tab:performance} compares the performance of SHD classification running on Loihi~2.
We did not repeat this analysis for SSC as the results would be very similar.
On the Jetson Orin Nano, we measured the runtime of the model using CUDA events at batch size 1, for minimal latency and at batch sizes (2, 4, 8, 16, 32, 64, 128), to find the minimum energy per inference. Following \citet{shrestha2024efficient}, we measured power usage by adding the power values reported by \lstinline{jtop} on the \lstinline{CPU_GPU_CV} and \lstinline{SOC} rails.
We averaged these during an initial \SI{20}{\second} of idling and during the simulation to obtain `static' and `dynamic' power values.
Using these values and the previously calculated simulation times, we calculated the total and dynamic energy per sample and hence the Energy Delay Product of SHD classification.
None of these models fully occupy the Jetson Orin Nano's GPU when batch size is 1, so latency is largely limited by the overhead of launching three GPU kernels at each simulation timestep (around \SI{10}{\micro\second} of latency per kernel).
Therefore, with batch size 1, the energy and latency per sample is relatively constant across architectures, and with and without delays.
With non-delayed models, this latency can be amortised across large batches, allowing the energy per sample to be reduced by up to $20\times$.
We would expect that switching to \SI{8}{\bit} weights would improve performance in this scenario by around $4\times$  due to the reduction in memory bandwidth, but this is not currently supported by GeNN. 
However, the optimal batch size for models with heterogeneous delays is much smaller, meaning adding delays incurs around a $6\times$ increase in energy on the Jetson.
This is because, while GeNN accumulates all inputs from non-delayed synapses within GPU registers, \emph{delayed} inputs are accumulated in global memory.
For this accumulation operation to be efficient, the delay buffers must fit into the GPU's \SI{2}{\mebi\byte} L2 cache, severely reducing the maximum batch size.

We conducted energy consumption and processing time measurements using an Oheo Gulch single-chip Loihi 2 system. To ensure accurate steady-state measurements, we loaded a single input sample into on-chip memory and then executed multiple processing cycles (commonly referred to as IO-unconstrained mode~\citep{shrestha2024efficient}).
Similar to Jetson, all model architectures can be readily parallelized across the available neurocores.
On Loihi 2, we observe that the feedforward networks use about $30\%$ more energy than the recurrent networks. 
Such an energy difference is expected, as the feedforward networks have 2 hidden layers, whereas the recurrent networks have only 1, i.e., the feedforward networks have twice as many hidden neurons, and this corresponds to an increased computational burden.
Importantly, Loihi 2 supports performant synaptic delays, and indeed we see that synaptic delays only marginally increase energy in both recurrent and feedforward networks, accompanied by at most a small increase in latency.
\section{Conclusions and Further Work}

In this work, we enhanced our previous pipeline~\citep{shoesmith2025eventprop} for efficiently training SNNs using EventProp in mlGeNN and deploying them to the Loihi 2 neuromorphic system by adding support for synaptic delays. After quantization of weights to 8 bits, our models suffer minimal accuracy loss while delivering significant energy savings compared to an embedded GPU. 

Previous research has shown that combining delays with structural plasticity is especially promising~\citep{meszaros2024learning} and that delays become particularly valuable in sparse connectivity settings~\citep{hammouamrilearning} as well as in recurrent architectures with very small numbers of hidden neurons~\citep{meszaros2025efficient}. Therefore, important avenues for future work include exploring if sparsely connected models with delays continue to perform well in temporally more complex tasks and to explore the accuracy-energy trade-off of models with delays, sparse connectivity and varying numbers of hidden neurons on Loihi 2.

 62 delay steps seem to be a strong constraint, but when using \SI{10}{\milli\second} timesteps, 30 delay steps seem to be optimal~\citep{hammouamrilearning}. In this work, we conducted our timestep hyperparameter search up to a maximum of \SI{2}{\milli\second}, but \SI{1}{\milli\second} timesteps resulted in the best performing models. The learning method relies on continuous dynamics, which seems to be sensitive to bigger timesteps. Tasks where precise temporal information over long periods of time is crucial could be an area where our methods are the better option. 
 
 Previous work achieved high accuracy in keyword recognition using more complex neuron models but without any delays~\citep{baronig2024advancing,higuchibalanced}, hinting that there are other ways to improve the computational capabilities of SNNs. Due to Loihi's programmable neuromorphic cores, such neuron models can be implemented effectivity~\citep{frady2022efficient} and exploring them in combination with delays is an interesting and a relatively underexplored direction~\citep{deckers2024co}.
 \balance
\printbibliography

\end{document}